\documentclass{ieeeaccess}
\usepackage{cite}
\usepackage{amsmath,amssymb,amsfonts}
\usepackage{algorithmic}
\usepackage{graphicx}
\usepackage{textcomp}
\usepackage{tabularx}

\usepackage{caption}
\usepackage{subcaption}
\usepackage{makecell}

\usepackage{hyperref}
\usepackage{float}

\def\BibTeX{{\rm B\kern-.05em{\sc i\kern-.025em b}\kern-.08em
    T\kern-.1667em\lower.7ex\hbox{E}\kern-.125emX}}
\begin{document}
\history{Date of publication xxxx 00, 0000, date of current version xxxx 00, 0000.}
\doi{10.1109/ACCESS.2017.DOI}

\title{Lidar Annotation Is All You Need}
\author{\uppercase{Dinar Sharafutdinov}\authorrefmark{1},
\uppercase{Stanislav Kuskov\authorrefmark{2}, \uppercase{Saian Protasov}\authorrefmark{3}, and Alexey Voropaev}.\authorrefmark{4}}
\address[1]{Evocargo LLC, Moscow, Russia, (e-mail: dinar.sharafutdinov@evocargo.com)}
\address[2]{Evocargo LLC, Moscow, Russia, (e-mail: stanislav.kuskov@evocargo.com)}
\address[3]{Evocargo LLC, Moscow, Russia, (e-mail: sayan.protasov@evocargo.com)}
\address[4]{Evocargo LLC, Moscow, Russia, (e-mail: alx.voropaev@gmail.com)}
\tfootnote{This work was supported by EVOCARGO LLC.}

\markboth
{Sharafutdinov \headeretal: Lidar Annotation Is All You Need}
{Sharafutdinov \headeretal: Lidar Annotation Is All You Need}

\corresp{Corresponding author: Dinar Sharafutdinov (e-mail: d.sharafutdinov@yahoo.com).}

\begin{abstract}

In recent years, computer vision has transformed fields such as medical imaging, object recognition, and geospatial analytics. One of the fundamental tasks in computer vision is semantic image segmentation, which is vital for precise object delineation. Autonomous driving represents one of the key areas where computer vision algorithms are applied. The task of road surface segmentation is crucial in self-driving systems, but it requires a labor-intensive annotation process in several data domains. The work described in this paper aims to improve the efficiency of  image segmentation using a convolutional neural network in a multi-sensor setup. This  approach leverages lidar (Light Detection and Ranging) annotations to directly train image segmentation models on RGB images. Lidar supplements the images by emitting laser pulses and measuring reflections to provide depth information. However, lidar’s sparse point clouds often create difficulties for accurate object segmentation. Segmentation of point clouds requires time-consuming preliminary data preparation and a large amount of computational resources. The key innovation of our approach is the masked loss, addressing sparse ground-truth masks from lidar point clouds. By calculating loss exclusively where lidar points exist, the model learns road segmentation on images by using lidar points as ground truth. This approach allows for seamless blending of different ground-truth data types during model training. Experimental validation of the approach on benchmark datasets shows comparable performance to a high-quality image segmentation model. Incorporating lidar reduces the load on annotations and enables training of image-segmentation models without loss of segmentation quality. The methodology is tested by experiments with diverse datasets, both publicly available and proprietary. The strengths and weaknesses of the proposed method are also discussed. The work facilitates efficient use of point clouds for image model training by advancing neural network training for image segmentation.

\end{abstract}


\titlepgskip=-15pt

\maketitle

\section{Introduction}

Computer vision has made huge advances in recent years, revolutionizing domains such as autonomous vehicles, medical imaging, and object recognition. Image segmentation is one of the fundamental tasks in computer vision, which plays a crucial role in accurately identifying and delineating objects of interest within an image. Modern self-driving systems rely on image segmentation, using convolutional neural networks in their perception pipelines. The task of image segmentation is effectively addressed by current state-of-the art models. But it requires a substantial amount of data and costly annotation.

Most self-driving systems use sensors such as lidars (Light Detection and Ranging) for various computer vision tasks. Lidar technology is widely utilized for obtaining detailed depth information about a scene by emitting laser pulses and measuring their reflections. The technology has found many applications in fields such as robotics, self-driving cars, and environmental mapping. While lidar data provide accurate depth information, they are unable to precisely segment objects within the scene. Lidars typically generate sparse point clouds. They measure distance to the surfaces of objects but cannot match visible-spectrum cameras when it comes to providing complete information about the shape, texture, and color of objects. Also, lidar measurements have limited accuracy in certain situations; for example, when dealing with transparent or reflective surfaces where the reflected laser pulses may be distorted or absorbed. Object segmentation based on lidar data can be challenging because the points obtained are not always directly associated with specific objects. For instance, in dense clusters of objects or at road intersections, points from separate objects can be mixed within a single point cloud.

These inconveniences make it difficult to create high-quality datasets and to train models using lidar point clouds. Utilizing two data domains makes it possible to harness the advantages of each approach but this demands significant resources for annotating data. An effective fusion of point clouds and images in training 2D road segmentation has the potential to enhance prediction quality and reduce annotation costs.


In this work, we propose a novel approach that effectively leverages lidar annotations to train image segmentation models directly on RGB images. By using lidar annotations as ground truth labels, we can project the information they provide in order to guide the segmentation process. Instead of manually annotating each image, we use lidar annotations to create road area labels. This allows us to focus on the annotation of only one type of data. Subsequently, we train an image segmentation model on projected lidar data to obtain a model capable of automatically segmenting roads in images.

The proposed method offers several advantages. Firstly, only lidar annotations are required, which significantly reduces the annotation workload. Secondly, incorporating lidar data by mixing them with images improves segmentation quality. Thirdly, our approach is flexible and applicable to various types of images and segmentation tasks. It not only optimizes the annotation process but also enables fusion of lidar and RGB data. This helps segmentation models to handle challenging scenarios such as low-light conditions, complex backgrounds, and objects of various shapes more successfully. By mixing different types of data we reduce the number of images that are needed in order to obtain high-quality results from the training process.

We measure the efficiency of our approach by carrying out comprehensive experiments using benchmark datasets, comparing the performance of our method with a high-quality image segmentation model. The results show comparable accuracy and efficiency, which confirms the benefits of incorporating lidar data in the image segmentation model training process.

This paper offers a detailed description of our approach and presents experimental results that showcase the effectiveness and applicability of our method. We also compare it with traditional methods for training segmentation models and discuss the advantages and limitations of our approach. 

The contributions of the paper are as follows:

\begin{itemize}
\item We propose a novel flexible and effective approach to image segmentation using convolutional neural networks and projected lidar point-cloud data as a ground truth.

\item We evaluate the proposed method on several datasets, compare it with a model trained using standard 2D ground truth, and consider the specifics of our approach, including mixing capabilities and differences in sensor setup. The paper is supported by the published GitHub repository\footnote{ \href{https://github.com/Evocargo/Lidar-Annotation-is-All-You-Need}{https://github.com/Evocargo/Lidar-Annotation-is-All-You-Need}} with the method implementation, processed datasets, and code base for future research.
\end{itemize}

\section{Related work}

Road segmentation is a common task for autonomous vehicle software and an important part of any self-driving system. There are many approaches to segmentation of roads on image \cite{mendes2016exploiting} \cite{alvarez2012road} \cite{shinzato2012fast}. Currently, neural network-based semantic segmentation is the most commonly used solution for the road segmentation task. Semantic segmentation \cite{thoma2016survey} consists in clustering parts of an image that belong to the same class of objects. This is a form of pixel-level prediction because each pixel in an image is classified according to a category.

Classical models of semantic image segmentation consist of two main parts: an encoder and a decoder. The encoder reduces the spatial resolution of the input data to extract important features. The decoder makes an upsample representation of features into a full-size segmentation map \cite{sediqi2021novel}. In the task of semantic segmentation, ground truth segmentation masks are usually used for training. Such masks encode which category each pixel belongs to. There are many approaches to semantic image segmentation for road segmentation that use segmentation masks and require segmentation datasets to be annotated on images, such as U-Net \cite{ronneberger2015u} base implementations, PSPNet \cite{pspnet}, etc., which allow high-precision image clustering.

Lidars are also used in autonomous vehicles. They provide information about the environment where cameras fail, such as in low-light conditions. Lidar data \cite{li2020deep} is a cloud of points reflected from surfaces and read by lidar. A point cloud, as well as a camera image, can be used for segmentation task \cite{zhang2020polarnet} \cite{xu2103deep}, and it should be labeled, too. Segmentation on lidar data requires more computing resources and additional training operations. Lidar-based methods show good quality, but they have limitations as well \cite{yan20222dpass}. Point-based methods such as PointNet \cite{qi2017pointnet} are generally time-consuming. Projection-based methods \cite{chen2021rangeseg} are very efficient approaches for lidar point clouds, but the accuracy of segmentation that they can achieve is limited. Voxel-based methods are a compromise between speed and accuracy, but they are still less accurate than image segmentation methods. Another way is multirepresentation fusion methods such as LIF-Seg\cite{zhao2021lif}. These methods combine multiple representations (e.g., points, projection images, and voxels) and utilize feature fusion among different network branches, which requires more computing resources, and the annotation of images and annotation of lidar points are mutually dependent.

There is another group of approaches for lidar semantic segmentation that do not involve neural networks. Such methods as \cite{Chu2017AFG}, \cite{5164280}, and \cite{5979818} are usually applied in limited cases and require customization for each situation. Their inference speed is highly dependent on the amount of data. However, if we use homogeneous data, they allow getting precise road segmentation results. It makes this group of methods a possible option for road surface annotation for lidar data.

The complexity of training lidar-based models as well as the costliness of point cloud labeling lead to the idea of using point clouds in another representation \cite{chen2017multi} \cite{rangenet++}. SqueezeSeg \cite{squeezeseg} proposes converting point clouds to dense spherical view and then processing them as image input for neural network. In contrast, VolMap \cite{volmap} is an example of a semantic segmentation model for 360-degree lidar data in volumetric bird's-eye view representation. PolarNet \cite{zhang2020polarnet} proposes the use of a polar bird's-eye view representation of point clouds. Multi-Projection Fusion
(MPF) framework \cite{alnaggar2021multi} incorporates several representations of the point cloud to minimize loss of information and increase the performance of the model. These approaches outperform standard 3D deep neural networks and allow for the segmentation of objects on a lidar point cloud. At the same time, they usually neglect image data and make predictions in intermediate space (spherical, bird-eye view, polar).

Another group of methods fuses information from several sensors, such as a monocular camera and lidar. The main challenge with such approaches is that images and point clouds are in different data domains. PLARD \cite{chen2019progressive} adapts point clouds to images by projecting them and applying altitude difference-based transformation. Then, the feature space adaptation allows for the fusion of image features and lidar data features in the neural network. In another approach \cite{lv2018novel}, images and lidar data in the form of lidar grid maps are processed by separate streams of FCN. Then they are fused by the Fusion Layer, which allows getting road prediction in the bird's-eye view representation. The authors in \cite{gu20183} propose the fusion of two road segmentation results. One comes from an inverse-depth aware fully convolutional neural network (IDA-FCNN), which integrates images and reprojected point clouds. The second prediction was produced by the inverse-depth histogram-based line scanning method on lidar's 3D point cloud. An approach from \cite{boulch2017unstructured} proposes a conversion of a point cloud to a mesh and the use of several image views of it. These image views in the form of both RGB and depth data are then fused in different ways in several segmentation neural network architectures.

These approaches also show promising results in terms of the quality of their predictions. But they may require annotated data from several sensors, which could be difficult to get. Also, such approaches use complex network architectures with various processing and fusion steps. This could lead to unstable training, and it demands more computational resources.

In our approach, we decided to train an image-based semantic segmentation model with the help of masked loss and projected road ground truth points from lidar space. The latter could be obtained either from point cloud annotation or from the secondary lidar road segmentation model. Such an approach allows predicting road surfaces in the image space using the advantages of lidar data and, at the same time, without data annotation in several domains.

\section{Approach}

\begin{figure*}[h] 
    \centering
    \includegraphics[width=0.8\textwidth]{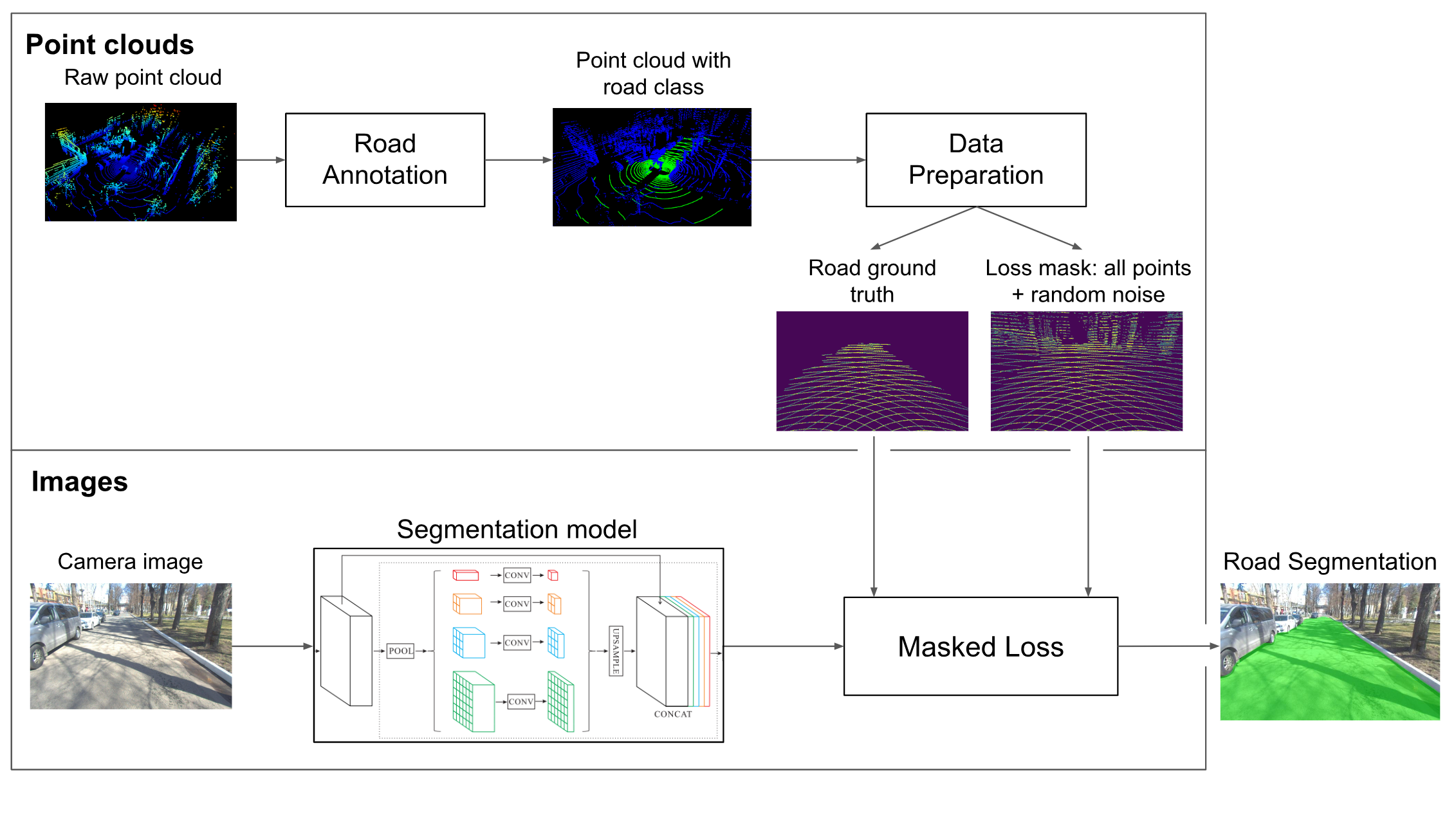}
    \caption{The overall scheme of our approach}
    \label{fig:scheme}
\end{figure*}

The overall pipeline of our approach is depicted in Fig. \ref{fig:scheme}. It consists of four main parts: point cloud road annotation, data preparation, masked loss, and the segmentation model itself. Firstly, we get the data with road annotations in the point cloud domain. After that, we project points using homogeneous transformations and camera parameters. Then, using the projected points we get the road ground truth and mask for loss calculation with added random noise. Images from the camera are processed by the segmentation model. Predictions and masks from the previous step are utilized by the Masked loss, which allows training the model using sparse ground truth data. Finally, after model training, we get an image with a segmented road. The training procedure as well as the Masked loss allow mixing projected ground truth with traditional 2D masks, which makes the approach flexible in terms of data. In the following subsections, we describe each part in detail.

\subsection{Point cloud road annotation}
To use lidar data during training, we need road surface annotation on point clouds. This could be done either manually using a point cloud annotation tool, such as SUSTechPOINTS \cite{SUSTech}, or using classical approaches for road surface detection \cite{Chu2017AFG}, \cite{5164280}, \cite{5979818}, which were mentioned in the Related work section. Depending on the data, road annotation could be done relatively fast thanks to the homogeneous nature of the road surface. A classical approach could allow getting road annotation without manual annotation at all, but it needs fine-tuning for specific data.

\subsection{Data preparation}
An obtained point cloud with road annotation is projected on the image plane using homogeneous transformations. For such projection, we use synchronized camera and lidar frames, accompanied by camera parameters, and a transformation matrix from lidar to camera frame. To project homogeneous point $ \mathbf x = (x, y, z, 1)^T$ in lidar frame coordinates to point $ \mathbf y = (u, v, 1)^T$ on an image plane, we use the equation:

\begin{equation}
\mathbf y = \mathbf K \mathbf T ^ {camera} _ {lidar} \mathbf x 
\end{equation}

where $ \mathbf K $ is the camera intrinsic matrix and $ \mathbf T ^ {camera} _ {lidar} $ is the transformation matrix to transform lidar coordinate to camera coordinate.

After transformation, we get points on the image as a mask of pixels, both for road class and for all the other points from the lidar scan. We use points in the form of two masks. The first one is the road mask (Fig. \ref{reproj_road}). In contrast with the standard segmentation training scheme, the mask here is sparse, so it couldn't be used in such a form directly. To overcome this issue, we use Masked loss, which will be explained in the next subsection. Additionally, we need all the other points from a lidar scan, so the second mask contains an entire lidar scan, including the road surface. Since the lidar points are mainly located in the lower part of the image, the upper part of the image does not have any points at all. It could lead to inaccurate predictions in that area. Because of that, we add random points in the upper half of the mask (negative class) to balance the distribution of the points where we will calculate the loss. The union of these two masks will be used as a space for loss calculation (Fig. \ref{loss_mask}).

\begin{figure}
     \centering
     \begin{subfigure}[b]{0.235\textwidth}
         \centering
         \includegraphics[width=\textwidth]{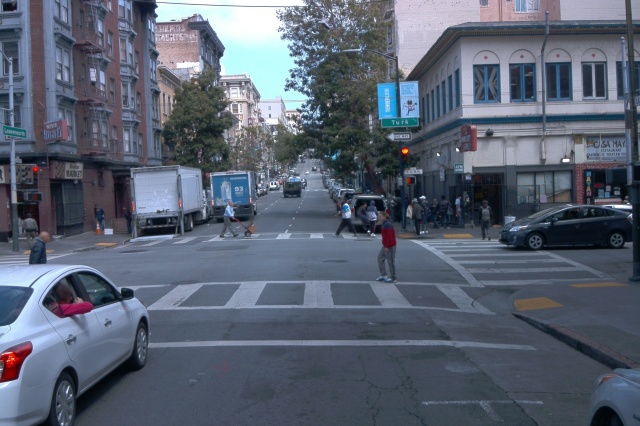}
         \caption{Camera image}
         \label{normal_img}
     \end{subfigure}
     \hfill
     \begin{subfigure}[b]{0.235\textwidth}
         \centering
         \includegraphics[width=\textwidth]{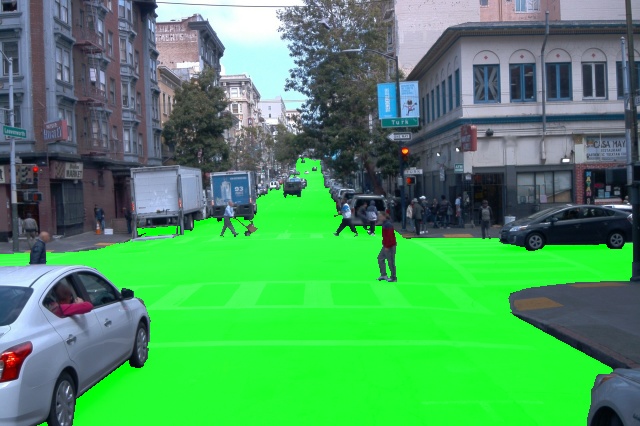}
         \caption{2D road mask}
         \label{2d_road_mask}
     \end{subfigure}
     \hfill
     \begin{subfigure}[b]{0.235\textwidth}
         \centering
         \includegraphics[width=\textwidth]{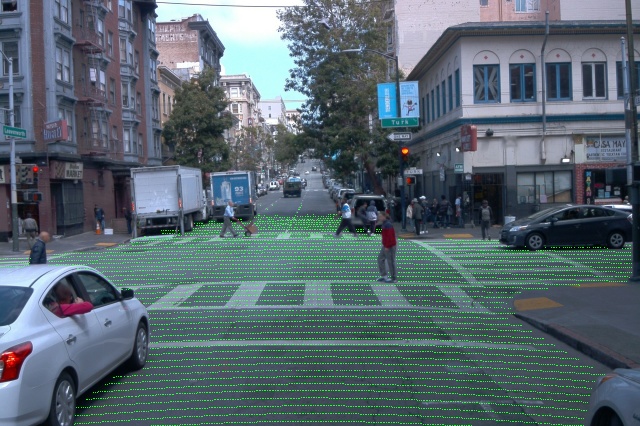}
         \caption{Projected road points}
         \label{reproj_road}
     \end{subfigure}
    \hfill
     \begin{subfigure}[b]{0.235\textwidth}
         \centering
         \includegraphics[width=\textwidth]{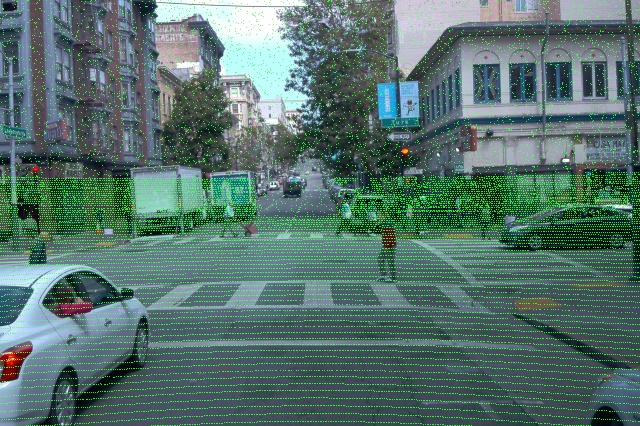}
         \caption{Loss calculation mask}
         \label{loss_mask}
     \end{subfigure}
        \caption{Examples of images and masks used during model training}
        \label{data_preparation}
\end{figure}

\subsection{Masked loss}
\begin{figure}[h!] 
    \centering
    \includegraphics[width=0.4\textwidth]{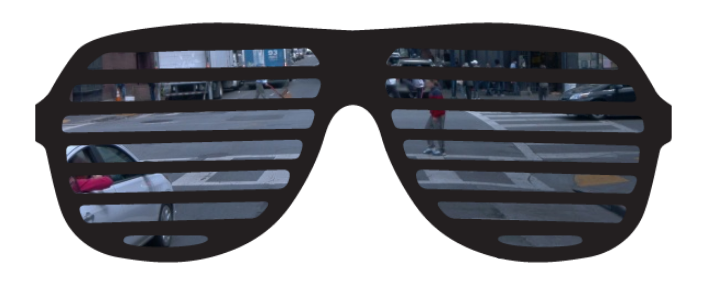}
    \caption{Shutter glasses view of the street}
    \label{shutter_glasses}
\end{figure}

To tackle the problem with the sparsity of road ground truth masks obtained from the lidar point clouds, we use Masked loss. The idea is to change the space of the loss calculation. Instead of using the whole image, we calculate the loss only in the pixels where we have lidar points, which are all the projected points. In other words, we force the model to learn road segmentation by measuring the error of predictions on a grid of points. It could be compared to looking at the image using shutter glasses (shutter shades). Even though they have lines, which might decrease visibility, the human eye can see almost the same picture. A schematic illustration of such a view is shown in Fig. \ref{shutter_glasses}. In the same way, the model gets enough information about the scene by calculating loss in specific areas of the image, which still covers most of the important parts including the road surface. Masked loss $L_{MASKED}$ for each image could be formulated in such a way:

\begin{equation}
L_{MASKED} = - \frac{1}{M} \sum_{i=1}^{N} m_i (y_i\log(p_i) + (1 - y_i)\log(1 - p_i)) 
\end{equation}

where $y_i$ indicates whether pixel class is actually the road or not, $p_i$ is a model prediction of pixel class, $m_i$ is an indicator showing whether there is a projected point in this pixel, $N$ is a total amount of pixels in the image, and $M$ is a total amount of lidar points projected on the image, e.g., all non-zero values in $m$.

In essence, it is a binary cross-entropy loss with an applied per-pixel mask, averaged by the number of lidar points projected on the image. For images with a standard mask of 2D road ground truth, we use a mask of ones for each pixel on the image. In this case, $m$ is an array of ones, and $N=M$. Which is the same as using non-masked binary cross-entropy loss. This makes it possible to mix two types of ground truth data during the training of the model.

\subsection{Model}
As a model, we use PSPNet semantic segmentation model \cite{pspnet}, which is well-studied by the computer vision community. PSPNet and its variations show good results for segmentation tasks in different domains as well as for road segmentation\cite{pspnet_compare} \cite{ge2022pfanet}. Moreover, it has benchmarks on various datasets, which are useful for research purposes. RGB camera images are used as input to the model.

\section{Experiments and Discussion}
\label{sec:guidelines}

\subsection{Setup}
\textbf{Datasets.}
To validate the method presented in our article, we used two open-source datasets and one proprietary dataset.

Perception Waymo Open Dataset \cite{Sun_2020_CVPR} (version 1.4.0) was used for the first set of experiments. It contains synchronized image and lidar data from urban environments. Moreover, there are semantic segmentation annotations for both 3D and 2D data. This makes it convenient to run experiments in different data setups: 2D masks only, projected masks only, and a mix of these two types of masks. Unfortunately, 2D and 3D annotations are created separately and not for the entirely same images. Because of that, we filtered the dataset so that all the images have both 2D road masks and projected road masks ("Waymo with intersection" dataset). This allows a fair comparison of metrics in different setups. We got 1852 images in the train set and 315 images in the validation set after filtering. All the filtering code as well as filtered data are available on the github page of the project. Projection of points from point clouds was done using Waymo tools from the dataset repository. Moreover, we extracted all the images and points without intersection ("Waymo full" dataset). This dataset contains 12291 masks with 2D ground truth, 23691 masks with projected points, 35982 masks for mixing case, and 34130 images (the overall number of images is smaller because of the intersection between 2D masks and masks with projected points).

Another open-source dataset for training and inference was the KITTI-360 dataset \cite{kitti360}. KITTI-360 is a suburban driving dataset that includes richer input capabilities, comprehensive semantic instance annotations, and precise localization, facilitating research at the intersection of vision, graphics, and robotics. Just like the Waymo Open Dataset, this dataset contains the necessary data, such as lidar scans synchronized with camera images, semantic segmentation for 3D lidar scans (for training), and semantic segmentation for 2D images (for validation and mix training). After extracting labeled images that correspond to the projected labeled lidar scans, we got 49004 samples for training and 12276 samples for validation.

\begin{figure}[h!] 
    \centering
    \includegraphics[width=0.48\textwidth]{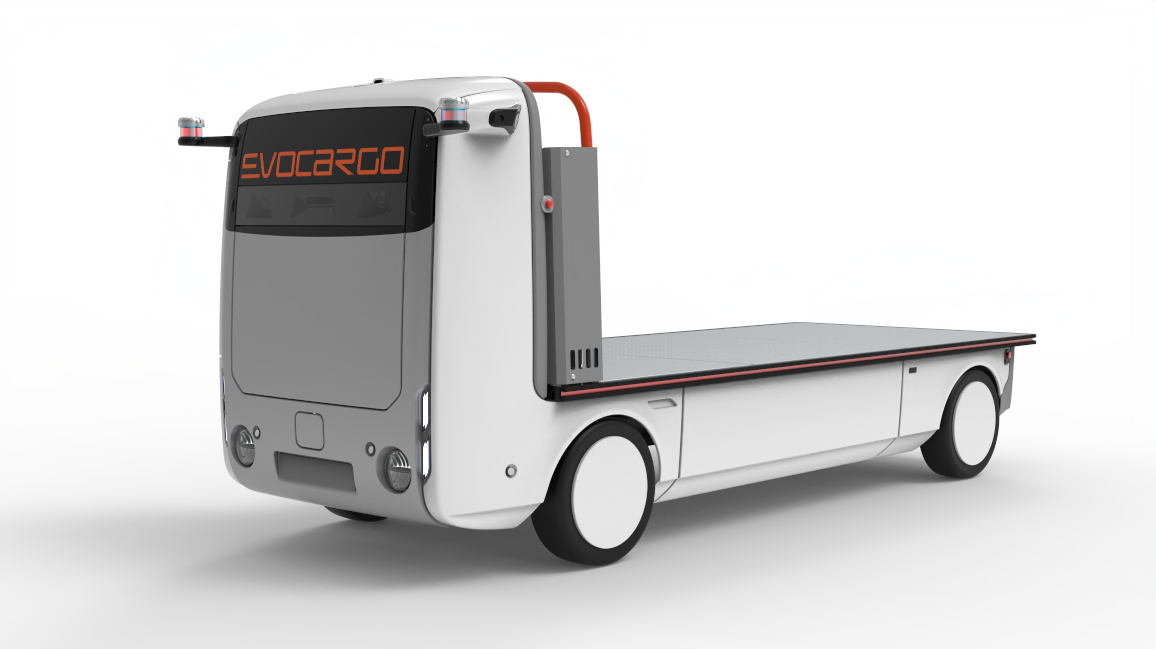}
    \caption{Evocargo autonomous vehicle used to collect proprietary data}
    \label{evo1}
\end{figure}

A proprietary dataset was collected using the autonomous cargo vehicle developed by Evocargo (Fig. \ref{evo1}). The vehicle is equipped with two Robosense RS-Helios 32 lidars and one Lucid Triton TRI023S-CC color camera. Camera images and lidar point clouds are precisely synchronized. Point clouds were annotated manually in the SUSTechPOINTS annotation tool \cite{SUSTech}. Using camera parameters and transforms, we projected point clouds on the image in the same way as in the previous experiments. The final dataset consists of 5000 images.

\textbf{Evaluation.}
Intersection-over-Union ($\mathbf{IoU}$) metric was used during training and inference stages: 

\begin{equation}
\mathbf{IoU} = \dfrac{|\mathbf P \cap \mathbf G |}{|\mathbf P \cup \mathbf G |}
\end{equation}

where $P$ and $G$ class predictions and ground truth, respectively.

\textbf{Implementation details.}
PyTorch was used as a deep learning framework. The PSPNet model implementation was taken from segmentation\_models.pytorch \cite{Iakubovskii} package. The YOLOP package \cite{wu2022yolop} was used as a code base.

For all experiments, we used the Adam optimizer with a momentum equal to 0.937. LambdaLR scheduler was applied for learning rate change during the training process. A standard set of augmentations such as perspective transformations, HSV color change, and left-right flips were used during training.

For experiments with Waymo with an intersection dataset, we trained the model for 300 epochs and used a learning rate equal to 0.001, with a final learning rate equal to 0.0005. Since the "Waymo full" is larger, we trained the model for 200 epochs and used a learning rate equal to 0.0001, with a final learning rate equal to 0.00005.

For the KITTI-360 and proprietary dataset experiments, we trained the model for 200 epochs (and 100 epochs for proprietary dataset respectively) and used the same learning rate as for the Waymo experiments, equal to 0.001, with a changing learning rate by the end of training equal to 0.0005.

\subsection{Results}

\textbf{Waymo dataset.}
We have conducted a set of three experiments: using only 2D road ground truth, only projected points as ground truth, and a mix of these types of ground truth. Moreover, we have trained and evaluated models on two types of data: "Waymo with intersection" and "Waymo full". A detailed explanation of dataset parsing could be found in Section III. A. The resulting IOU scores could be found in Table \ref{table:waymo1} and Table \ref{table:waymo2}.

From Table \ref{table:waymo1} we can conclude that the model trained on the small version of the dataset using only 2D ground truth has the baseline IOU at 96.1\%. Whereas for the model, which used only projected points, the score is equal to 94.4\%. It is still a high-quality result, but it is lower by 1.7\%. Experiment with mixed data shows 95.3\% IOU, which tells us that mixing two types of ground truth allows to increase quality compared to projected points only experiment and at the same time decreases quality compared to 2D masks only experiment.

By looking at the results on the "Waymo full" dataset in Table \ref{table:waymo2}, we can observe the decrease of quality on more difficult and variable data. The result produced by the model trained using only 2D data is 0.5\% lower for "Waymo full" than for "Waymo with intersection". At the same time, the result shown by the model trained using projected points is only 0.2\% lower. This means that the model trained using the projected points is more robust to variability of data and needs less data to get high quality results. This is also true for the model trained on a large dataset (-0.4\% and -0.1\% respectively). Also, there is a growth in quality for the mixing experiment of "Waymo full" (96.3\% vs. 96.1\% for the 2D only experiment), which means that the addition of projected data could increase the quality of the model trained only on 2D ground truth.
 
With the use of a larger dataset, we see the same gap between the 2D masks experiment and the projected points experiment. At the same time, the quality of the mixing experiment is 0.2\% higher than the 2D mask-only experiment. Moreover, the quality of both 2D-only and projected points only experiments increased compared to the "Waymo with intersection" dataset.

Also, we can notice that there is no drastic change in prediction quality between models trained on small and large versions of datasets, especially taking into account huge differences in version sizes. The reason for that is that data in the large version of the dataset has a larger frame rate, but scenes are similar to the ones in the small version. Only details and dynamic objects could make a difference in such a case.

\begin{table}[h]
\begin{center}
\centering
\caption{Road segmentation results (\% of IoU) on validation split of "Waymo with intersection".}
\label{table:waymo1}
\begin{tabular}{|l|c|c|}
\hline \bfseries Experiment & \multicolumn{2}{c|}{\bfseries Trained on} \\
\hline & \bfseries Waymo with intersection & \bfseries Waymo full \\ 

\hline 2D only (baseline) & 96.1 & 96.5 \\ 
\hline Projected 3D only & 94.4 & 94.8 \\ 
\hline mix 2D + projected 3D & 95.3 & 96.5 \\ 
\hline
\end{tabular}
\end{center}
\end{table}

\begin{table}[h]
\begin{center}
\centering
\caption{Road segmentation results (\% of IoU) on validation split of "Waymo full".}
\label{table:waymo2}
\begin{tabular}{|l|c|c|}
\hline \bfseries Experiment & \multicolumn{2}{c|}{\bfseries Trained on} \\
\hline & \bfseries Waymo with intersection & \bfseries Waymo full\\ 

\hline 2D only (baseline) & 95.6 & 96.1 \\ 
\hline Projected 3D only & 94.2 & 94.7 \\ 
\hline mix 2D + projected 3D & 95.1 & 96.3 \\ 
\hline
\end{tabular}
\end{center}
\end{table}

\textbf{KITTI-360 dataset.}
Three distinct experimental conditions were systematically applied to the KITTI-360 dataset: the utilization of only 2D road ground truth, the exclusive use of projected points as ground truth, and a composite approach integrating both types of ground truth. Contrary to the experimental procedures implemented for the Waymo dataset, no partitioning between comprehensive and partial datasets was undertaken for KITTI-360, given the inherent consistency of data correspondence within the KITTI-360 repository.

Our experiments show that the use of projected points for training continues to demonstrate satisfactory results performance, as shown in Table \ref{table:kitti1}. However, when we compare this with the results from the Waymo dataset, we observe a noticeable IOU deviation, with a decline of 2.7\% from the baseline. This shift in values can be linked with the unique features of the KITTI-360 dataset. A significant observation from our study is the difference in point distribution on the KITTI-360 imagery as compared to the Waymo dataset, evident from Fig. \ref{img:lidar_gt}. The dispersion of these points likely correlates with factors such as the number of laser beams in a lidar system and the total count of lidars in the setup. A greater number of points typically leads to a denser distribution. Despite these nuances, the approach of utilizing only points for mask training remains relatively consistent, with minor variances observed.

As for training the model on the mixed data between re-projected 3D points and 2D image masks, the results are similar to the results of the corresponding experiment on the "Waymo full" dataset. Mixing data increases IOU compared to task baseline by +0.4\%.

\begin{table}[h]
\centering
\caption{Road segmentation results (\% of IoU) on validation split of KITTI-360 dataset.}
\label{table:kitti1}
\begin{tabularx}{0.5\textwidth}{|X|c|}
\hline 
\bfseries Experiment & \bfseries KITTI-360 \\
\hline 
2D only (baseline) & 92.3 \\ 
\hline 
Projected 3D only & 89.6 \\ 
\hline 
mix 2D + projected 3D & 92.7 \\ 
\hline
\end{tabularx}
\end{table}

\begin{figure}
     \centering
     \begin{subfigure}[b]{0.235\textwidth}
         \centering
         \includegraphics[width=\textwidth]{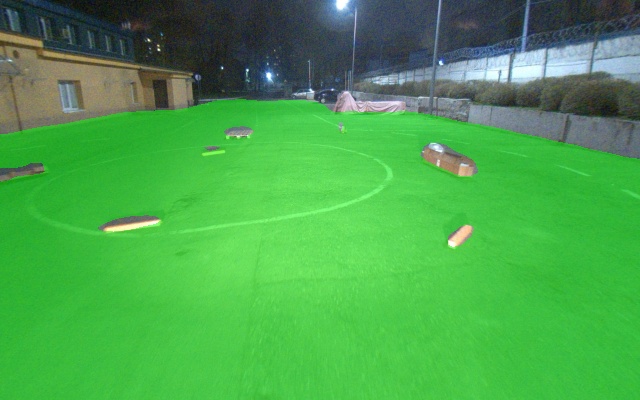}
         \caption{Obstacles}
         \label{normal_img}
     \end{subfigure}
     \hfill
     \begin{subfigure}[b]{0.235\textwidth}
         \centering
         \includegraphics[width=\textwidth]{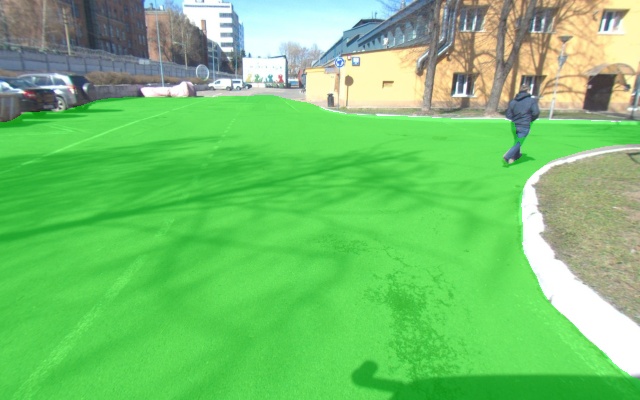}
         \caption{Person}
         \label{2d_road_mask}
     \end{subfigure}
     \hfill
     \begin{subfigure}[b]{0.235\textwidth}
         \centering
         \includegraphics[width=\textwidth]{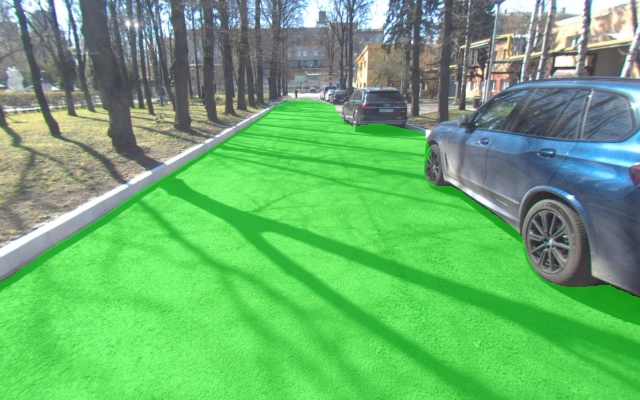}
         \caption{Car object}
         \label{reproj_road}
     \end{subfigure}
    \hfill
     \begin{subfigure}[b]{0.235\textwidth}
         \centering
         \includegraphics[width=\textwidth]{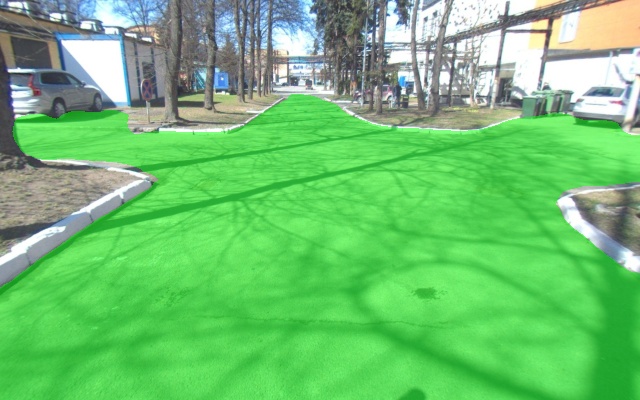}
         \caption{Empty road}
         \label{loss_mask}
     \end{subfigure}
        \caption{Road prediction examples of the model on proprietary data}
        \label{xxinference_our_data}
\label{fig:inference_our_data}
\end{figure}

\textbf{Proprietary dataset}
We have trained the same model on the dataset collected using the Evocargo autonomous vehicle. The only experimental setup here is the use of solely projected 3D points as a ground truth. The final predictions can be seen in Fig. \ref{fig:inference_our_data}. It could be noticed that despite differences in lidar points distribution compared to Waymo and Kitti setups, which could be seen in Fig. \ref{img:lidar_gt} (different density, distance between lidar rings, and shape of lidar rings), the model still trains well. Predictions for different types of scenes (e.g., with obstacles on the road, a walking person, parked cars, and a crossroad) are very precise, which means that this approach can be applied to different sensor setups and proprietary datasets.

\subsection{Discussion}

By examining the predictions in Fig. (\ref{img:waymo_infer_results}), we can make several observations. First of all, the predictions of the 2D-only experiment (the first row) and the projected 3D-only experiment (the second row) are very similar in terms of road coverage and quality of segmentation, which is confirmed by the metrics. Secondly, we could notice several limitations of using lidar-based ground truth, which is a limited ground truth distance and subsequent lower quality of predictions for distant points, as well as less precise predictions at object boundaries. Thirdly, we see how the mixing experiment (the third row) eliminates both the imperfections of the 2D ground truth-based model and the limitations of the lidar point-based model, which leads to better road segmentation.

\def\pagepart{0.24}
\def\imagewidth{1.0}
\begin{figure*}[htbp]
    \centering
    \begin{minipage}{\pagepart\textwidth}
        \centering
        \includegraphics[width=\imagewidth\linewidth]{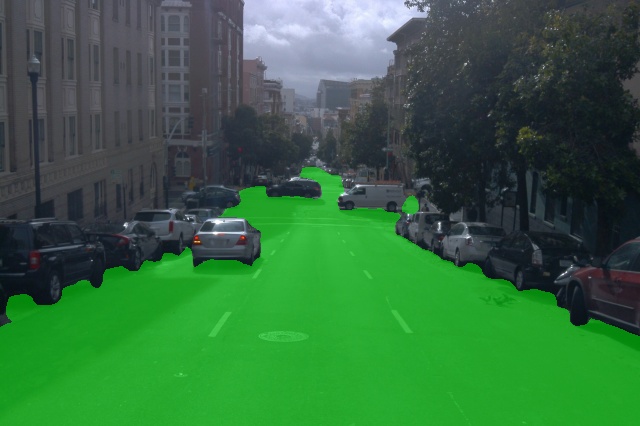}
    \end{minipage}
    \begin{minipage}{\pagepart\textwidth}
        \centering
        \includegraphics[width=\imagewidth\linewidth]{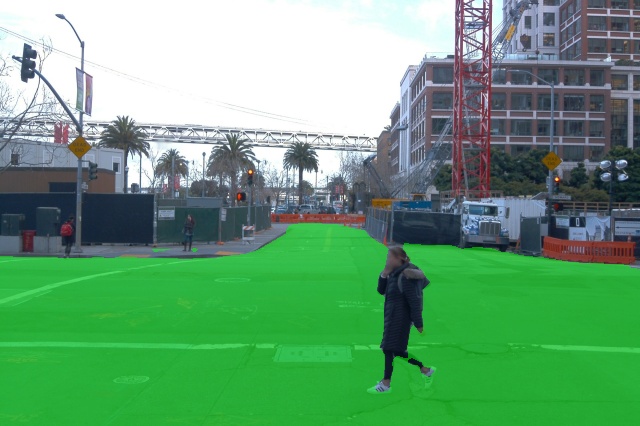}
    \end{minipage}
    \begin{minipage}{\pagepart\textwidth}
        \centering
        \includegraphics[width=\imagewidth\linewidth]{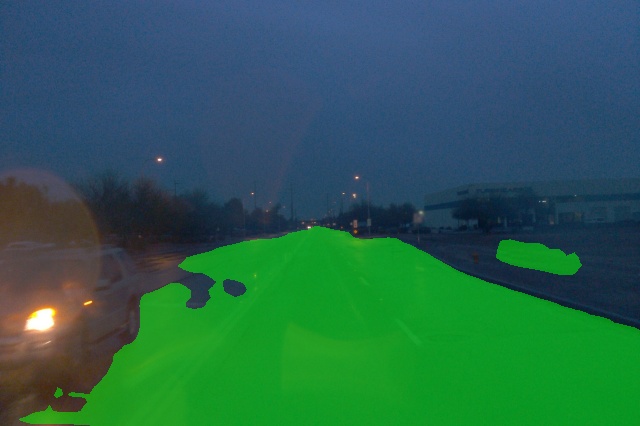}
    \end{minipage}
    \begin{minipage}{\pagepart\textwidth}
        \centering
        \includegraphics[width=\imagewidth\linewidth]{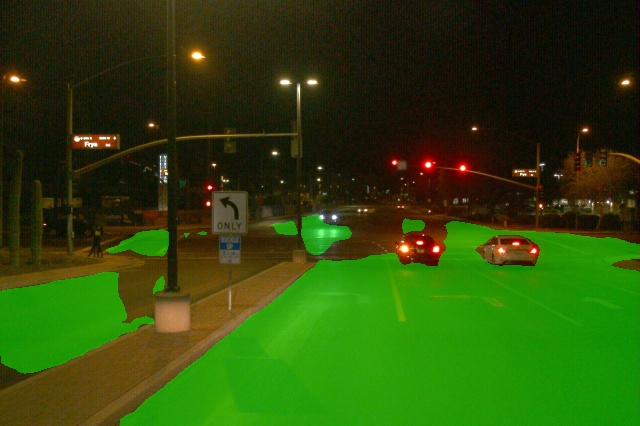}
    \end{minipage}
    
    \vspace{0.2cm} 
    
    \begin{minipage}{\pagepart\textwidth}
        \centering
        \includegraphics[width=\imagewidth\linewidth]{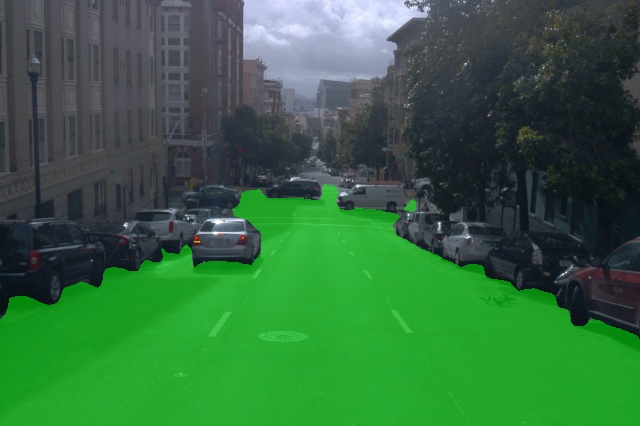}
    \end{minipage}
    \begin{minipage}{\pagepart\textwidth}
        \centering
        \includegraphics[width=\imagewidth\linewidth]{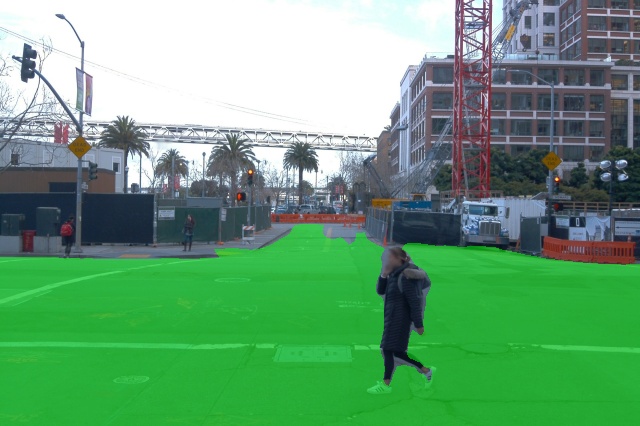}
    \end{minipage}
    \begin{minipage}{\pagepart\textwidth}
        \centering
        \includegraphics[width=\imagewidth\linewidth]{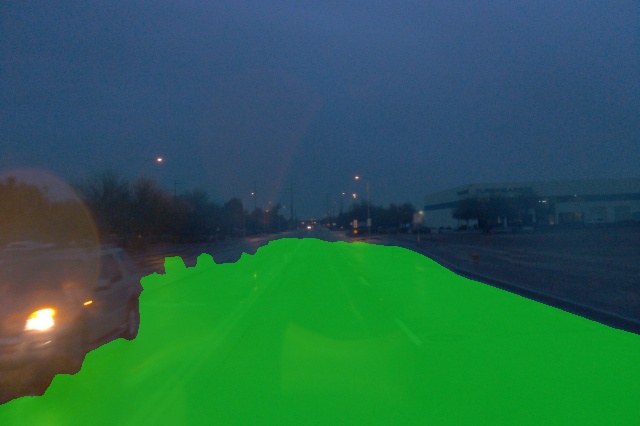}
    \end{minipage}
    \begin{minipage}{\pagepart\textwidth}
        \centering
        \includegraphics[width=\imagewidth\linewidth]{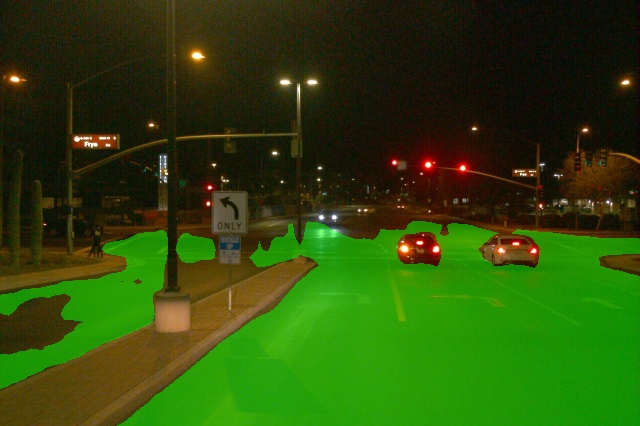}
    \end{minipage}

    \vspace{0.2cm} 
    
    \begin{minipage}{\pagepart\textwidth}
        \centering
        \includegraphics[width=\imagewidth\linewidth]{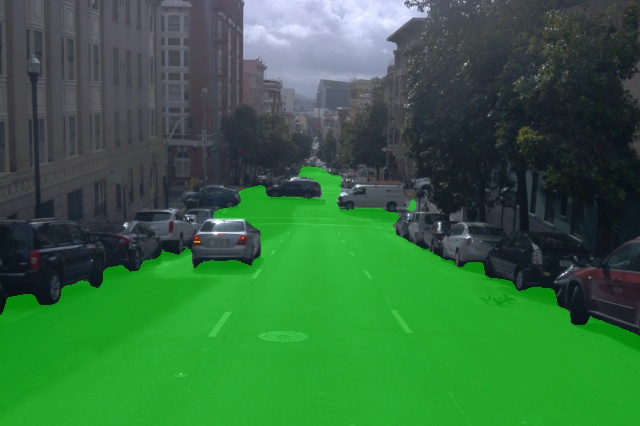}
    \end{minipage}
    \begin{minipage}{\pagepart\textwidth}
        \centering
        \includegraphics[width=\imagewidth\linewidth]{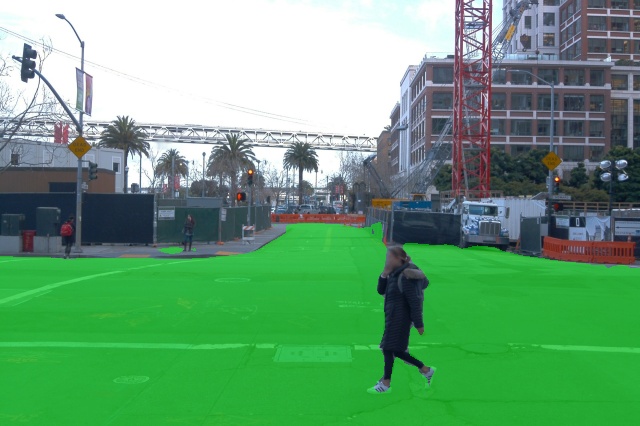}
    \end{minipage}
    \begin{minipage}{\pagepart\textwidth}
        \centering
        \includegraphics[width=\imagewidth\linewidth]{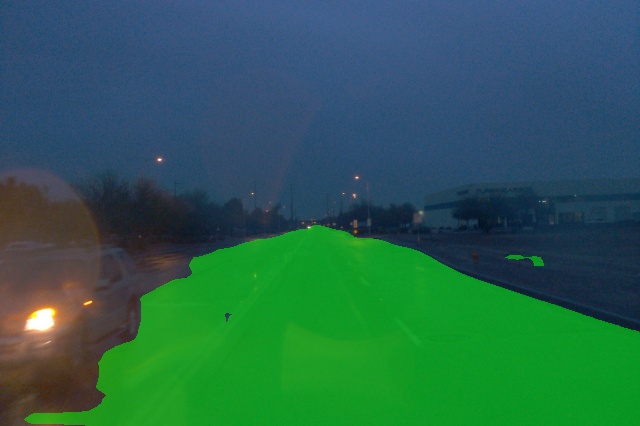}
    \end{minipage}
    \begin{minipage}{\pagepart\textwidth}
        \centering
        \includegraphics[width=\imagewidth\linewidth]{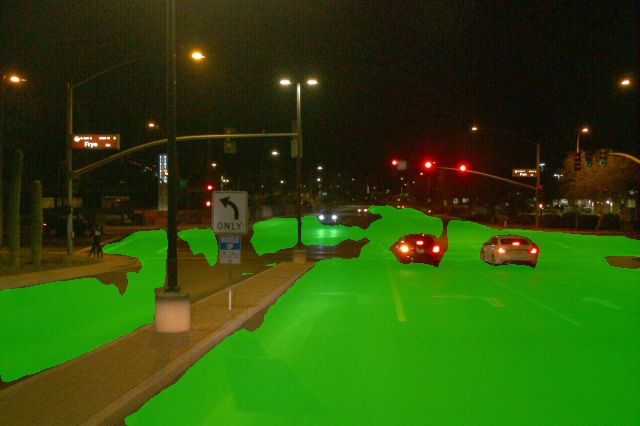}
    \end{minipage}

    \caption{Comparison of predictions of three models on the Waymo Open Dataset. From top to bottom: 2D only, projected 3D only, mix 2D + projected 3D}
    \label{img:waymo_infer_results}
\end{figure*}

\def\pagepart{0.48}
\def\imagewidth{1.0}
\begin{figure*}[htbp]
    \centering
    \begin{minipage}{\pagepart\textwidth}
        \centering
        \includegraphics[width=\imagewidth\linewidth]{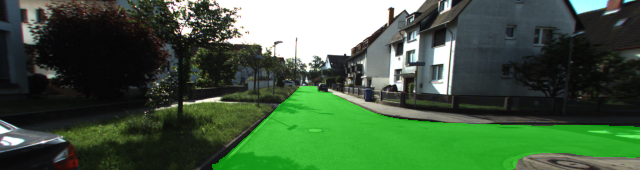}
    \end{minipage}
    \begin{minipage}{\pagepart\textwidth}
        \centering
        \includegraphics[width=\imagewidth\linewidth]{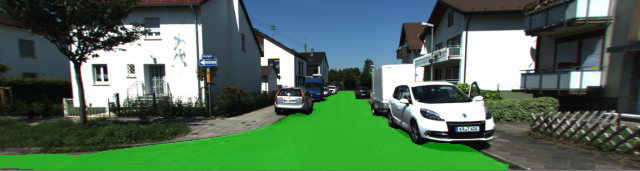}
    \end{minipage}
    
    \vspace{0.2cm} 
    
    \begin{minipage}{\pagepart\textwidth}
        \centering
        \includegraphics[width=\imagewidth\linewidth]{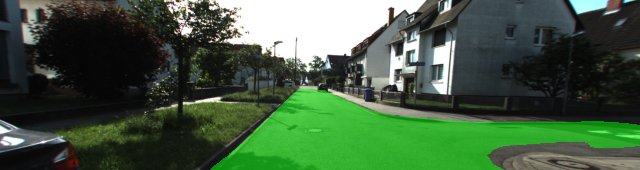}
    \end{minipage}
    \begin{minipage}{\pagepart\textwidth}
        \centering
        \includegraphics[width=\imagewidth\linewidth]{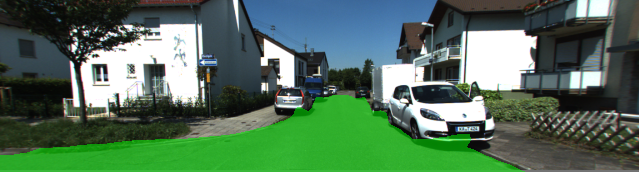}
    \end{minipage}

    \vspace{0.2cm} 
    
    \begin{minipage}{\pagepart\textwidth}
        \centering
        \includegraphics[width=\imagewidth\linewidth]{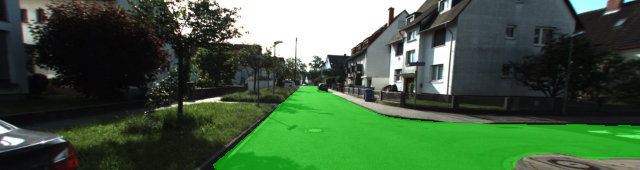}
    \end{minipage}
    \begin{minipage}{\pagepart\textwidth}
        \centering
        \includegraphics[width=\imagewidth\linewidth]{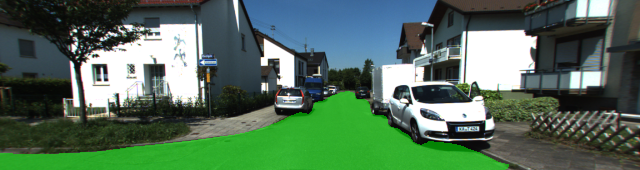}
    \end{minipage}

    \caption{Comparison of predictions of three models on KITTI-360 dataset. From top to bottom: 2D only, projected 3D only, mix 2D + projected 3D}
    \label{img:kitti_infer_results}
\end{figure*}

Based on that, we could conclude that the nature of lidar data influences the quality of image segmentation predictions. Since lidar has a limited range and receives data at a certain frequency, points on the image will also have a fixed range as well as less distinct boundaries of objects compared to 2D ground truth. This is the possible reason for the quality decrease of up to 2.5\%. At the same time, we could notice that the quality of prediction is very high, despite this peculiarity of lidar-based ground truth data. 

\begin{figure}[H] 
    \centering
    \includegraphics[width=0.48\textwidth]{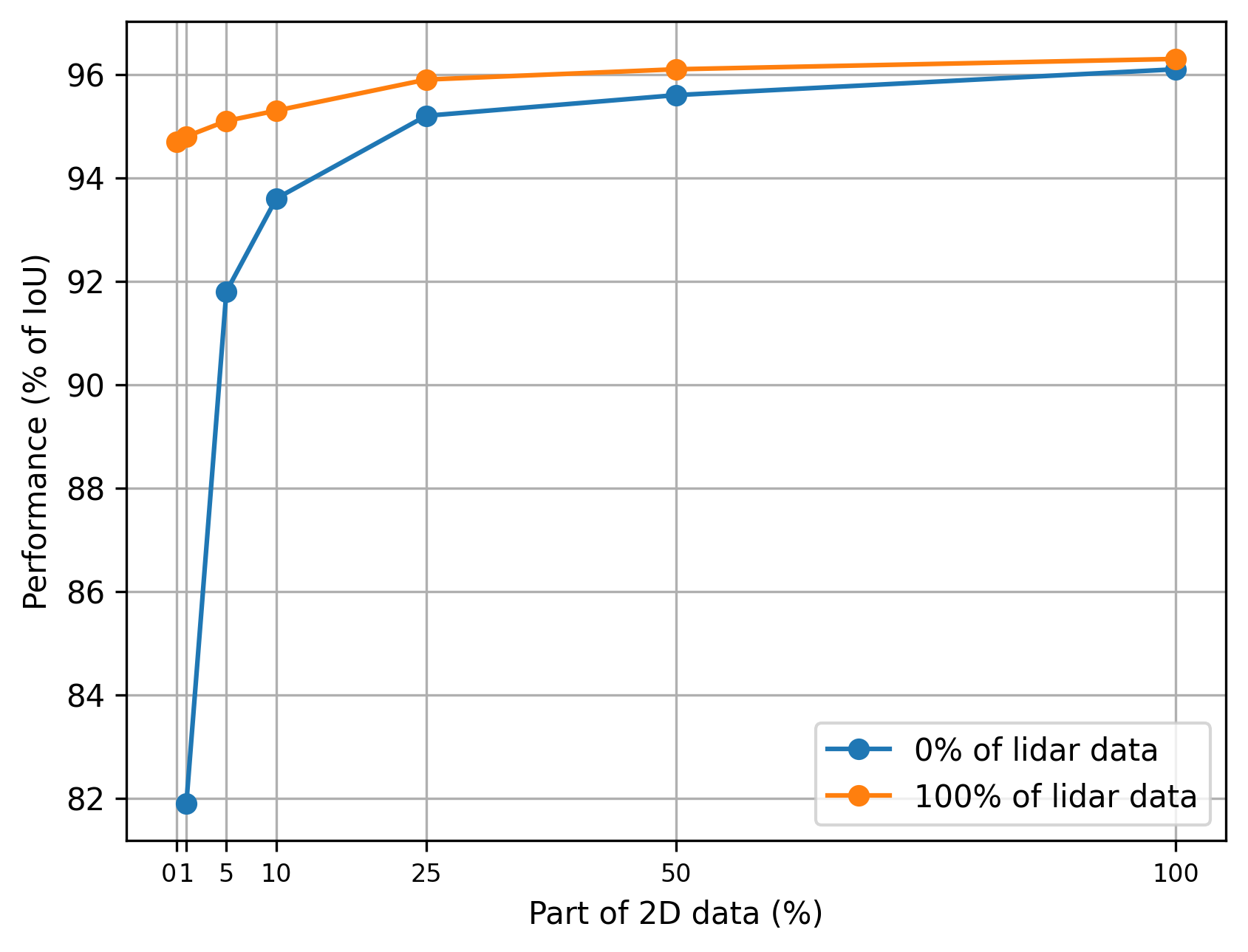}
    \caption{Road segmentation results (\% of IoU) on the validation split of the "Waymo full" dataset for models trained on different ratios of 2D image masks and lidar-based ground truth.}
    \label{mixing_plot}
\end{figure}

To explore the possibility of leveling out this feature, we conducted an experiment on different ratios of data mixing between 2D image masks and lidar-based ground truth. As we can see from the Fig. \ref{mixing_plot}, addition of 2D image masks to 100\% of lidar-based ground truth increases the quality of prediction. Use of 25\% of 2D image masks allows getting almost the same result as using 100\% of images solely (95.9\% vs. 96.1\%), whereas addition of 50\% of 2D image masks increases quality to 96.1\%. Moreover, the addition of 100\% of 2D image masks allows getting the highest overall performance, as shown in the previous section.

Regarding the experiments with the KITTI-360 dataset (Fig. \ref{img:kitti_infer_results}), the results appear similar to those from the Waymo dataset experiment. At first glance, the road segmentation seems very akin to the road segmentation trained solely on 2D road masks. As previously discussed, the metric decrease in this case is slightly higher than in the Waymo experiment (a decline of -2.7\%). We posit that the poorer metrics can be attributed to various factors, including hardware-related issues. The final results can be influenced by different densities of point projections onto the image, the image's perspective, the resulting mask of points projected onto the image, and the lidar-camera calibration. An intriguing observation from the KITTI-360 dataset is that after projecting points onto the image, the lidar appears to "look" beneath vehicles. This phenomenon is due to the placement of the lidar and the accessibility of those areas to the device's lasers (Fig. \ref{img:lidar_gt}), and these areas are considered during training.

%

\section{Conclusions}

In this paper, we have introduced a novel approach to road surface segmentation, leveraging both annotated lidar point clouds and traditional 2D image masks. Through our pipeline, which includes point-cloud road annotations, ground-truth data preparation, a segmentation neural network and a uniquely designed masked loss function, we showed how lidar-derived road masks can make neural networks perform better in an image segmentation task. Our approach uses fewer resources to annotate data from different types of sensors with comparable quality.

A notable advantage of our method is its flexibility. It can combine 2D image masks with 3D point-cloud masks in datasets during the training process. This versatility is very valuable, particularly when manual annotations for both 3D lidar points and images might be impractical (when the datasets are very large). The method was tested using two public datasets (Waymo Open Dataset and KITTI-360) and one proprietary dataset, and the results were promising. Performance of the model trained only on projected lidar point masks was slightly inferior to the 2D mask-trained model, but quality of the former was still high and was comparable to that of the latter. Performance metrics of the model trained on a combination of 2D masks and lidar projections matched or even surpassed the traditional 2D mask model, emphasizing the potential of this hybrid approach. Mixing experiments show that it is possible to reduce the amount of images needed for training by 50\% and to increase the amount of projected 3D data without losing quality of prediction.

Analysis of results from Waymo and KITTI-360 datasets, indicates that lidar characteristics, such as point distribution on images, range, and frequency, can influence segmentation outcomes. This variability suggests that, although lidar-based annotations are important, unique attributes of each dataset, including hardware specifics and environmental variables, should also be taken into account.

Future studies could  use different approaches. Enhancing the size of projected lidar points may improve coverage, but with some loss of accuracy. There is potential for expanding data-fusion techniques, harnessing multi-camera perspectives around a vehicle, and exploring diverse lidar setups. Understanding the nuances between different lidars is crucial, and there is a  need to test our approach in various conditions (winter and night-time scenarios, etc.). Finally, new neural architectures that incorporate projected lidar points as input require additional research, potentially optimizing the combination of 2D image annotations with 3D lidar data for enhanced segmentation.

\appendices

\def\pagepart{0.32}
\def\imagewidth{1.0}
\begin{figure*}[htbp]
    \centering
    \begin{minipage}{\pagepart\textwidth}
        \centering
        \includegraphics[width=\imagewidth\linewidth]{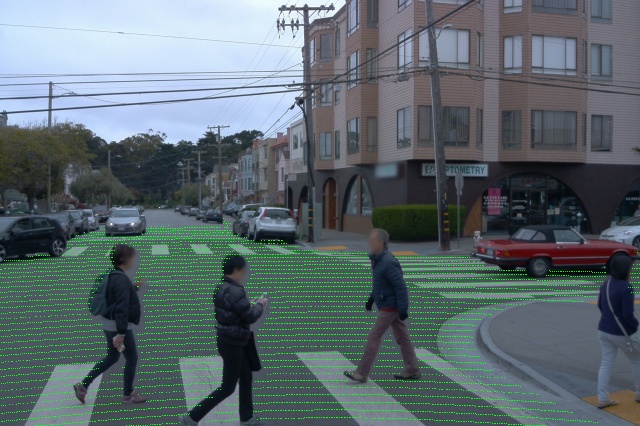}
    \end{minipage}
    \begin{minipage}{\pagepart\textwidth}
        \centering
        \includegraphics[width=\imagewidth\linewidth]{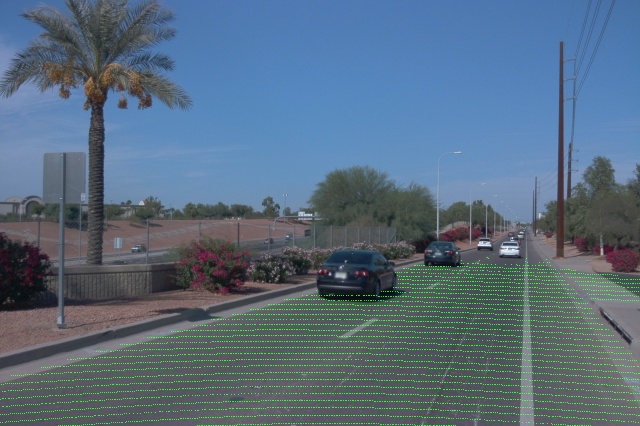}
    \end{minipage}
    \begin{minipage}{\pagepart\textwidth}
        \centering
        \includegraphics[width=\imagewidth\linewidth]{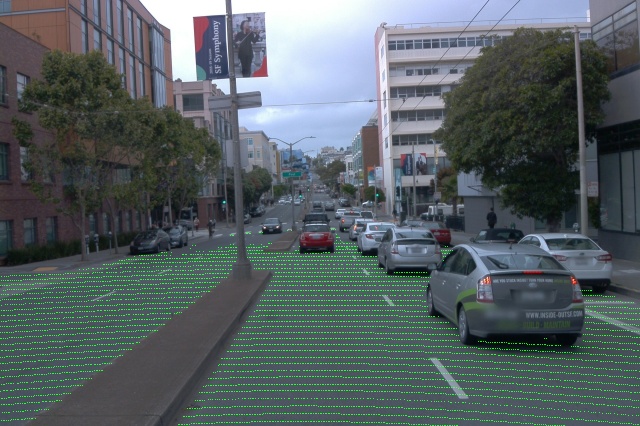}
    \end{minipage}
    
    \vspace{0.2cm} 

    \begin{minipage}{\pagepart\textwidth}
        \centering
        \includegraphics[width=\imagewidth\linewidth]{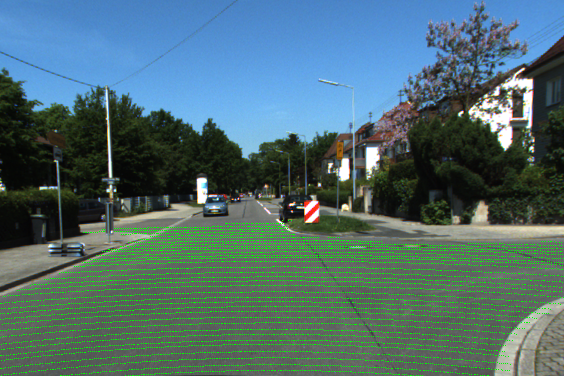}
    \end{minipage}
    \begin{minipage}{\pagepart\textwidth}
        \centering
        \includegraphics[width=\imagewidth\linewidth]{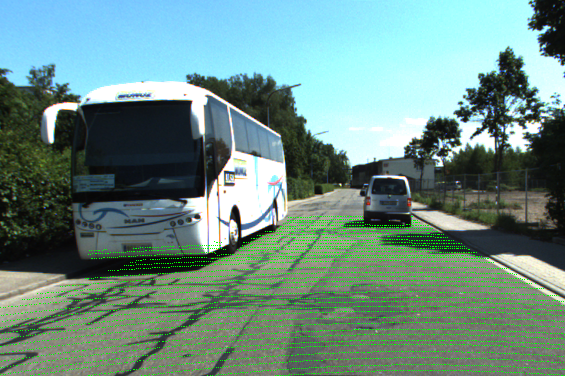}
    \end{minipage}
    \begin{minipage}{\pagepart\textwidth}
        \centering
        \includegraphics[width=\imagewidth\linewidth]{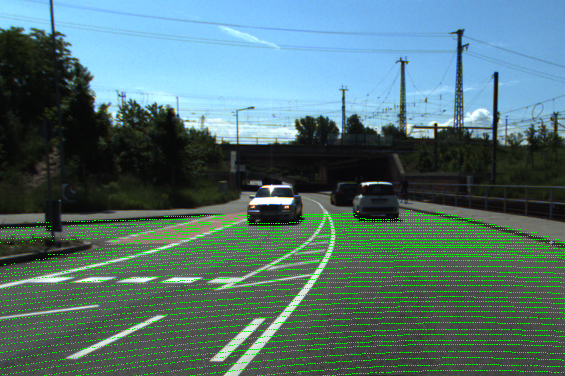}
    \end{minipage}


    \vspace{0.2cm} 
    
    \begin{minipage}{\pagepart\textwidth}
        \centering
        \includegraphics[width=\imagewidth\linewidth]{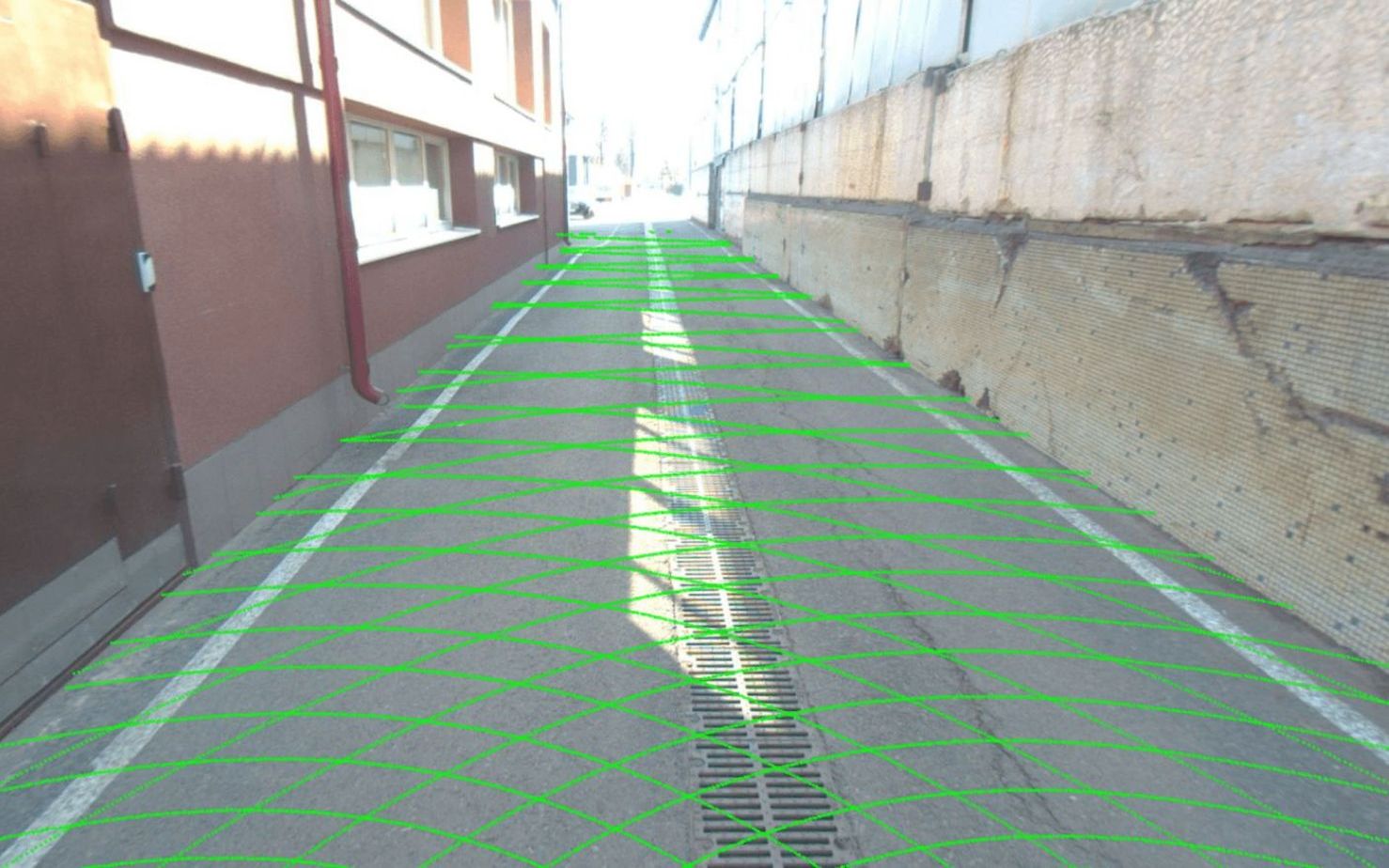}
    \end{minipage}
    \begin{minipage}{\pagepart\textwidth}
        \centering
        \includegraphics[width=\imagewidth\linewidth]{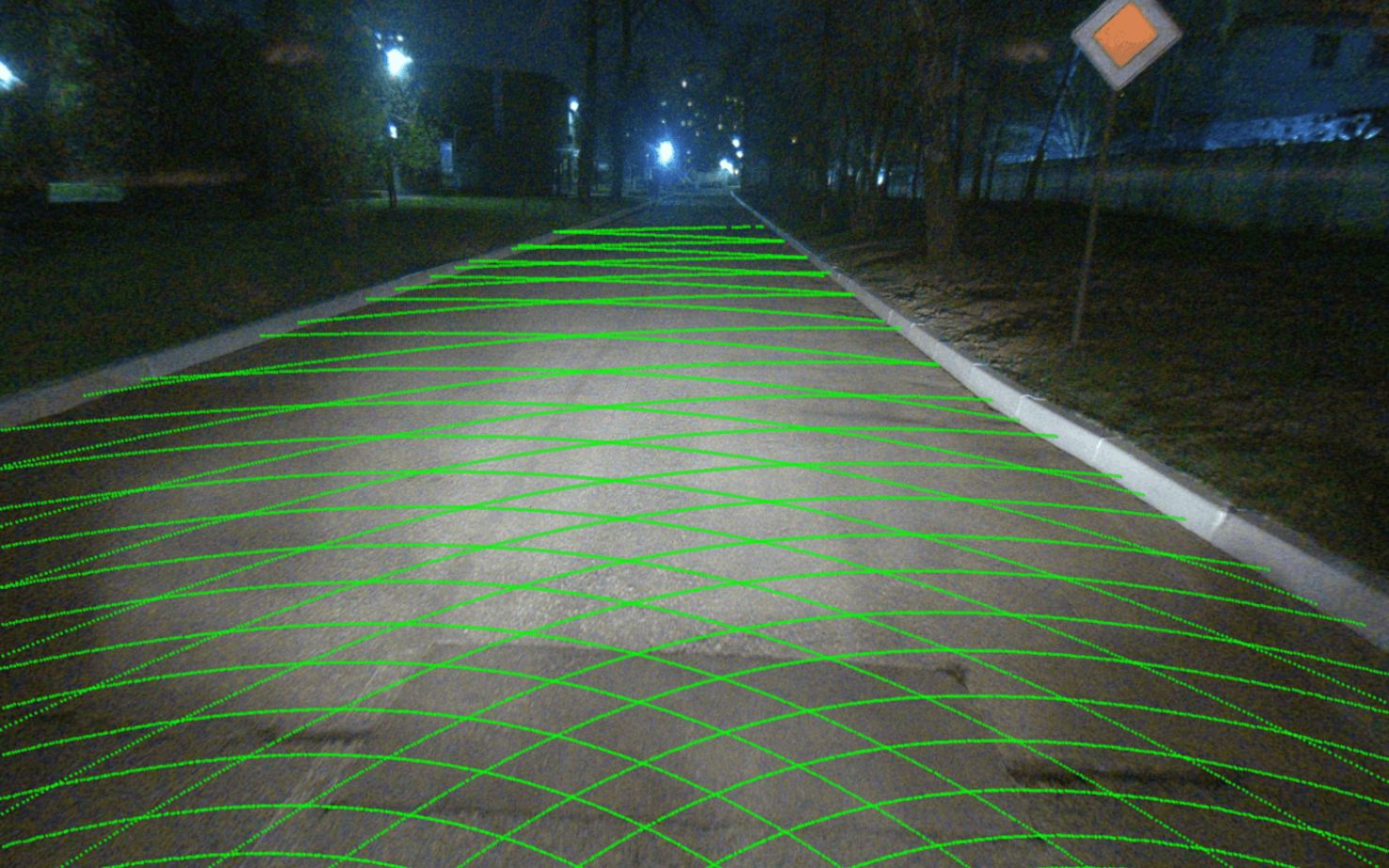}
    \end{minipage}
    \begin{minipage}{\pagepart\textwidth}
        \centering
        \includegraphics[width=\imagewidth\linewidth]{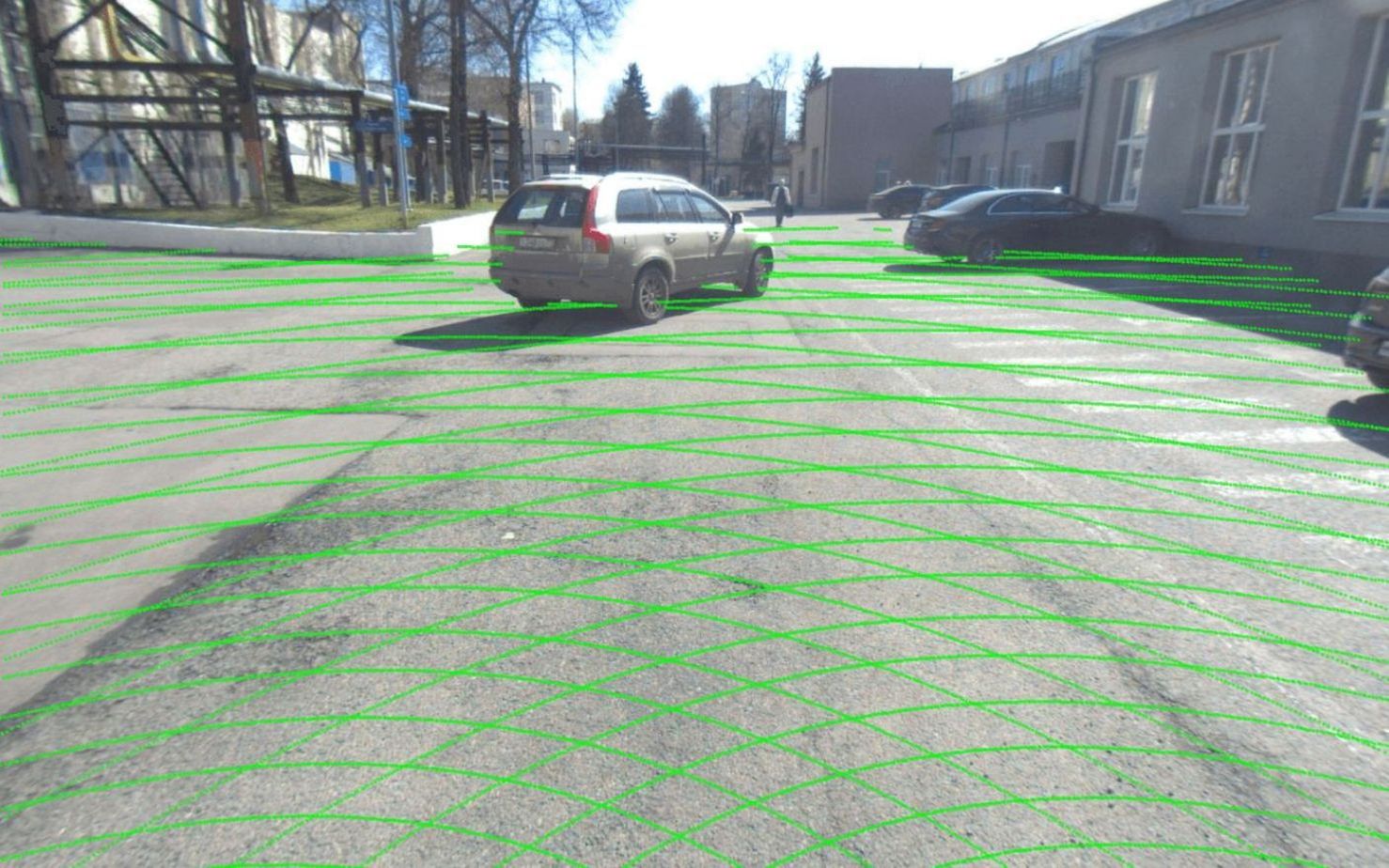}
    \end{minipage}


    \caption{Comparison of lidar-based road ground truth for three setups placed from top to bottom:
    Waymo Open Dataset (five proprietary lidar sensors\cite{waymo_open_dataset}), KITTI-360 (Velodyne HDL-64E lidar\cite{kitti360}), Proprietary dataset (two Robosense RS-Helios lidars)}
    \label{img:lidar_gt}
\end{figure*}

\bibliographystyle{IEEEtran}
\bibliography{references.bib}{}





\begin{IEEEbiography}
[{\includegraphics[width=1in,height=1.25in,clip,keepaspectratio]{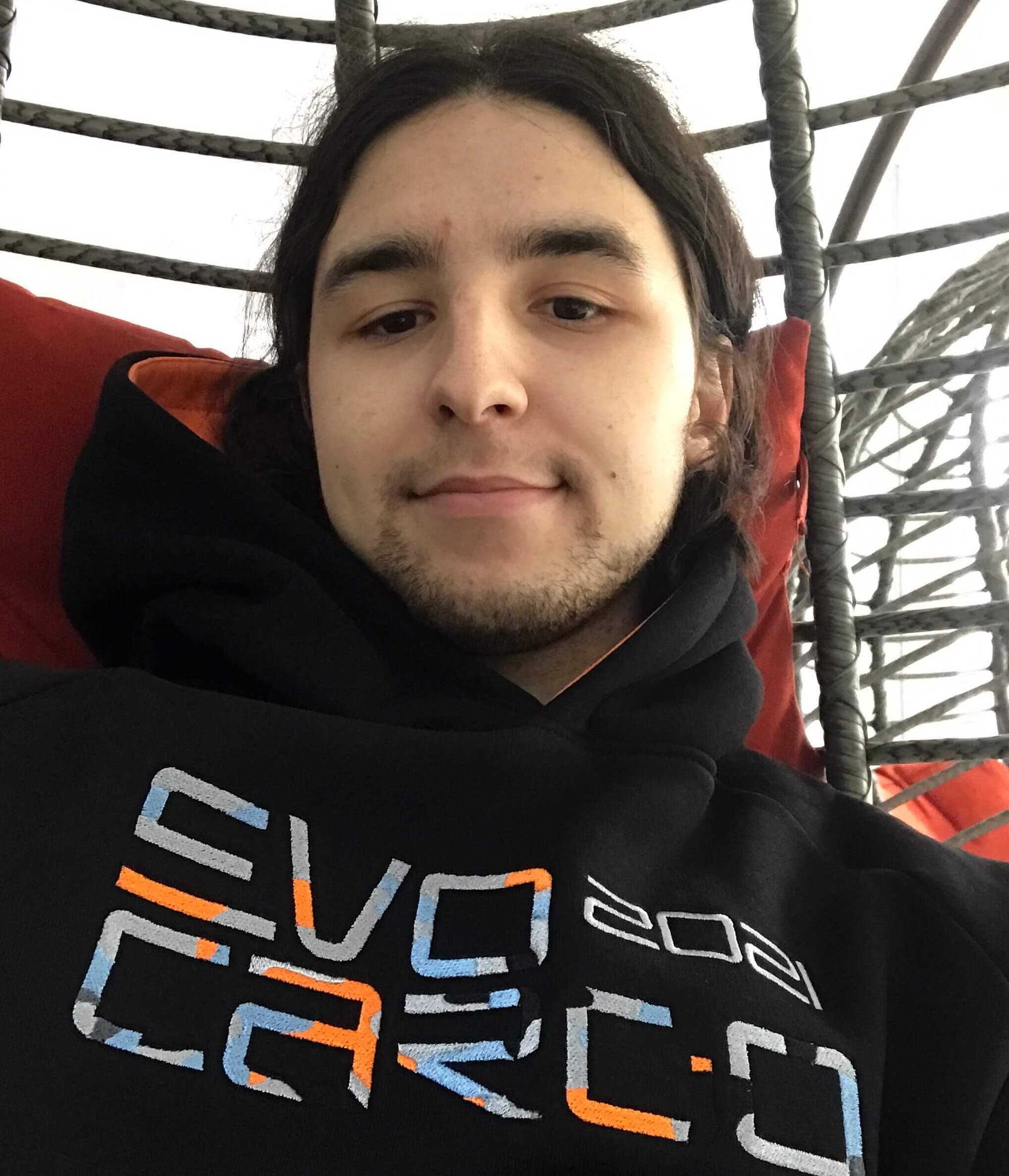}}]{Dinar Sharafutdinov} obtained a Master’s degree in Mathematics and Computer Science from Skolkovo Institute of Science and Technology, Moscow, Russia in 2021 and a Bachelor’s degree from Saint Petersburg State University, Saint Petersburg, Russia, in 2018. His research area includes computer vision, deep learning, generative models, autonomous robots, and self-driving technologies.

From 2019 to 2021, he was a member of the Mobile Robotics Lab under the supervision of Gonzalo Ferrer. He participated in various projects related to visual place recognition, unsupervised deep visual odometry, and point cloud fusion. Dinar worked on a project for evaluation of SLAM algorithms during a summer internship at Sber Robotics Lab (findings of the evaluation were published in a paper). Since 2021 he has worked as a computer vision engineer at Evocargo, Moscow, Russia, focusing on object detection and road segmentation models for self-driving vehicles used in autonomous logistics. He also works on multitask models, training pipelines, data pipelines, and research projects related to images and point-cloud fusion.
\end{IEEEbiography}

\begin{IEEEbiography}[{\includegraphics[width=1in,height=1.25in,clip,keepaspectratio]{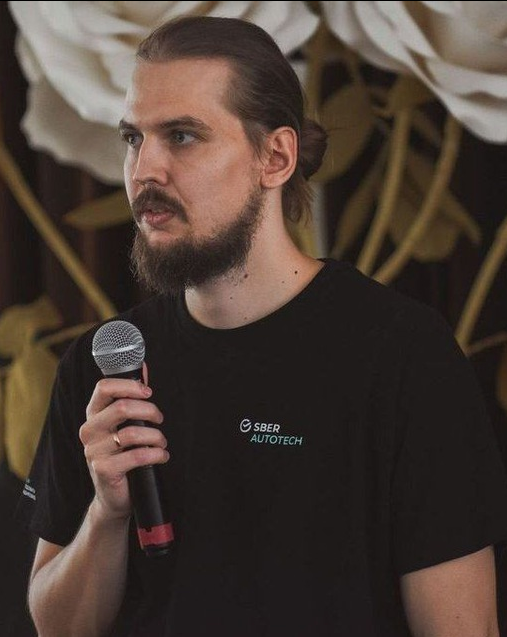}}]{Stanislav Kuskov} obtained a Bachelor of Science degree in Control and Data Management in Industrial Systems at Moscow State University of Mechanical Engineering, Moscow, Russia, in 2015 and a Master of Science degree in Computer Science from Moscow Polytechnic University, Moscow, Russia, in 2017. In 2016, he was a research assistant at Paderborn University, Paderborn, Germany. His research work has focused on the building of interface interactions between automatic differentiation libraries.

Stanislav worked as a software developer at Vist Robotics from September 2018 to July 2020, where he developed a computer vision system that enables self-driving trucks in coal mines to detect and avoid obstacles using security cameras. He also designed and implemented a number of other systems, including truck trajectory planning, wheel angle correction, braking control, functional testing, and a driver’s attention control system.

From November 2021 to April 2022, as tech lead at SberAutoTech, he worked on a multi-task architecture for dynamic object detection successfully optimizing and implementing such a system for self-driving vehicles. As a team lead at Evocargo since April 2022, Stanislav has been responsible for the development of computer vision system architectures and training pipelines for object detection and segmentation. He also leads a team focused on 2D tasks, including segmentation and detection and the application of various techniques to improve the quality of 2D algorithms.

\end{IEEEbiography}

\begin{IEEEbiography}[{\includegraphics[width=1in,height=1.25in,clip,keepaspectratio]{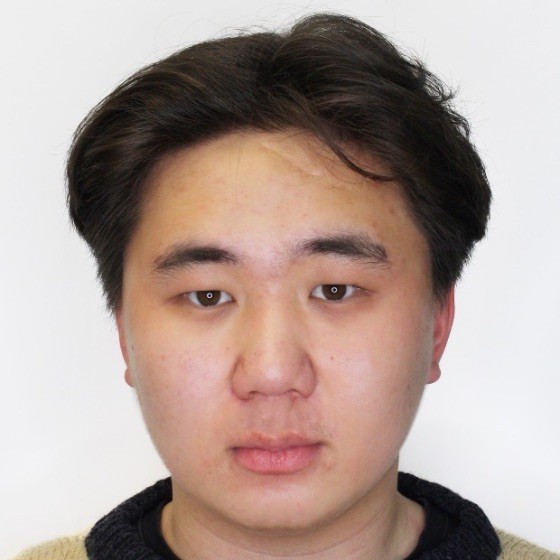}}]{Saian Protasov} obtained a Bachelor’s degree in physics from Novosibirsk State University, Novosibirsk, Russia, in 2019 and then a Master’s degree in Information Systems and Technologies from Skolkovo Institute of Science and Technology, Moscow, Russia, in 2021. His primary research interests are 3D computer vision, deep learning, autonomous mobile robotics, and self-driving technologies.

Between 2019 and 2021, he worked as a member of the team supervised by Dzmitry Tsetserukou at the Intelligent Space Robotics Laboratory. He was also part of the Skoltech Eurobot team, which achieved top placements in Russia and won accolades in the global Eurobot competition. His lab work has focused on advancing computer vision algorithms for mobile robots using deep learning approaches.

Since 2021 he has been affiliated with Evocargo LLC as a deep learning engineer, specialized in the development of 3D detection systems for the autopilot of self-driving vehicles.
\end{IEEEbiography}

\begin{IEEEbiography}
[{\includegraphics[width=1in,height=1.25in,clip,keepaspectratio]{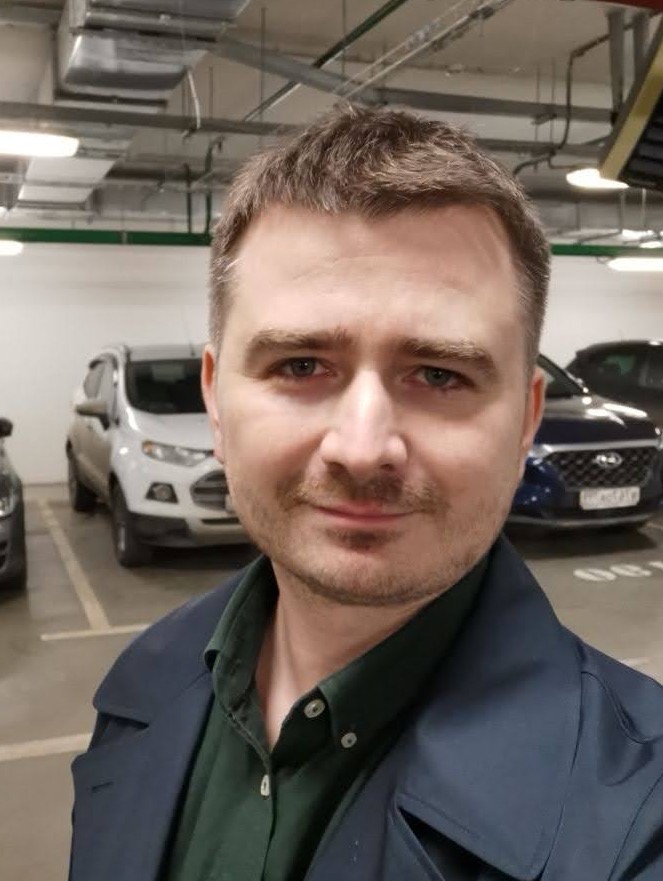}}]
{Alexey Voropaev} obtained a degree in software engineering from Pacific National University, Khabarovsk, Russia in Russia in 2007 and a Master of Science degree in computer science from Saarland University in Germany in 2009. His current research interests are autonomous driving software, computer vision, reinforcement learning, and high-performance computing.

From September 2009 to July 2021, Alexey served as a tech lead on various machine-learning and deep-learning projects at VK Group (formerly Mail.Ru), including speech recognition/synthesis, face recognition, and a range of image processing projects. From July 2021 to April 2022, he led a team of over 35 engineers developing a software stack for self-driving heavy-duty trucks, which were successfully tested on public roads. From April 2022 to July 2023, he oversaw a group of more than 15 engineers at Evocargo LLC working on the perception system for the N1 autonomous truck designed for operation at closed territories. The truck fleet is now operational at various production sites.
\end{IEEEbiography}

\EOD

\end{document}